\documentclass{article}

\PassOptionsToPackage{numbers, compress}{natbib}
\usepackage[final]{nips_2016}

\usepackage[utf8]{inputenc} 
\usepackage[T1]{fontenc}    
\usepackage{hyperref}       
\usepackage{url}            
\usepackage{booktabs}       
\usepackage{amsfonts}       
\usepackage{nicefrac}       
\usepackage{microtype}      
\usepackage{amsmath}        
\usepackage{color}
\usepackage{graphicx}
\usepackage{subcaption}

\title{Temporal Activity Detection in Untrimmed Videos with Recurrent Neural Networks}

\author{
    Alberto Montes \\
    ETSETB TelecomBCN\\
    Universitat Politècnica de Catalunya \\
    Barcelona, Catalonia/Spain \\
    \texttt{malberto@student.ethz.ch} \\
    \And
    Amaia Salvador \\
    Image Processing Group \\
    Universitat Politècnica de Catalunya \\
    Barcelona, Catalonia/Spain \\
    \texttt{amaia.salvador@upc.edu} \\
    \And
    Santiago Pascual \\
    TALP Research Center \\
    Universitat Politècnica de Catalunya \\
    Barcelona, Catalonia/Spain \\
    \texttt{santiago.pascual@tsc.upc.edu} \\
    \And
    Xavier Giro-i-Nieto \\
    Image Processing Group \\
    Universitat Politècnica de Catalunya \\
    Barcelona, Catalonia/Spain \\
    \texttt{xavier.giro@upc.edu} \\
}

\begin{document}

\maketitle

\begin{abstract}


    This work proposes a simple pipeline to classify and temporally localize activities in untrimmed videos. Our system uses features from a 3D Convolutional Neural Network (C3D) as input to train a a recurrent neural network (RNN) that learns to classify video clips of 16 frames. After clip prediction, we post-process the output of the RNN to assign a single activity label to each video, and determine the temporal boundaries of the activity within the video. We show how our system can achieve competitive results in both tasks with a simple architecture. We evaluate our method in the ActivityNet Challenge 2016, achieving a 0.5874 mAP and a 0.2237 mAP in the classification and detection tasks, respectively. Our code and models are publicly available at at: \url{https://github.com/imatge-upc/activitynet-2016-cvprw}

\end{abstract}

\section{Introduction}

Recognizing activities in videos has become a hot topic over the last years due to the continuous increase of video capturing devices and online repositories.
This large amount of data requires an automatic indexing to be accessed after capture.
The recent advances in video coding, storage and computational resources have boosted research in the field towards new and more efficient solutions for organizing and retrieving video content.

Impressive progress has been reported in the recent literature for video classification~\cite{tran2014learning,tran2015deep,wang2015towards,yao2015describing}, which requires to assign a label for the input video. While this task is already challenging, it has typically been explored with videos to be trimmed beforehand.
However, a video classification system should be able to recognize activities in untrimmed videos, and find the temporal segments in which they appear.
This second challenge has been recently proposed in the ActivityNet Challenge 2016~\cite{caba2015activitynet}, in which participants are asked to both provide a single activity for each video, as well as the temporal segment where the activity happened in the video.
In order to face both these challenges at the same time, we propose a simple pipeline composed of a 3D-CNN that exploits spatial and short temporal correlations, followed by a recurrent neural network which exploits long temporal correlations.


\section{Related work}


Several works in the literature have used 2D-CNNs to exploit the spatial correlations between frames of a video by combining their outputs using different strategies~\cite{gkioxari2015contextual,yeung2015end,ballas2015delving}. Others have tried using the optical flow as an additional input to the 2D-CNNN~\cite{wang2015towards}, which provides information of the temporal correlations. 

Later on, 3D-CNNs were proposed in~\cite{tran2014learning} (known as C3D), which were able to exploit short temporal correlations between frames and have demonstrated to work remarkably well for video classification~\cite{tran2014learning,tran2015deep}. C3D have also been used for temporal detection in~\cite{scnn_shou_wang_chang_cvpr16}, where multi-stage C3D architecture is used to classify video segment proposals.

For temporal activity detection, recent works have proposed the usage of Long Short-Term Memory units (LSTM)~\cite{hochreiter1997long}.
LSTMs are a type of RNNs that are able to better exploit long and short temporal correlations in sequences, which makes them suitable for video applications.
LSTMs have been used alongside CNNs for video classification~\cite{yao2015describing} and activity localization in videos~\cite{yeung2015every}.

In this paper, we combine the capabilities of both 3D-CNNs and RNNs into a single framework. This way, we design a simple network that takes a sequence of video features from the C3D model~\cite{tran2014learning} as input to a RNN and is able to classify each one of them into an activity category.


\section{Proposed Architecture}



We use the C3D model proposed in~\cite{tran2014learning} to extract features for all videos in the database. We split the videos in 16-frames clips and resize them to 171$\times$128 to fit the input of the C3D model. Features from the second fully connected layer (fc6) are extracted for each video clip.



\subsection{Architecture}
\label{sec:architecture}
\begin{figure}[ht]
\centering
\includegraphics[width=.5\linewidth]{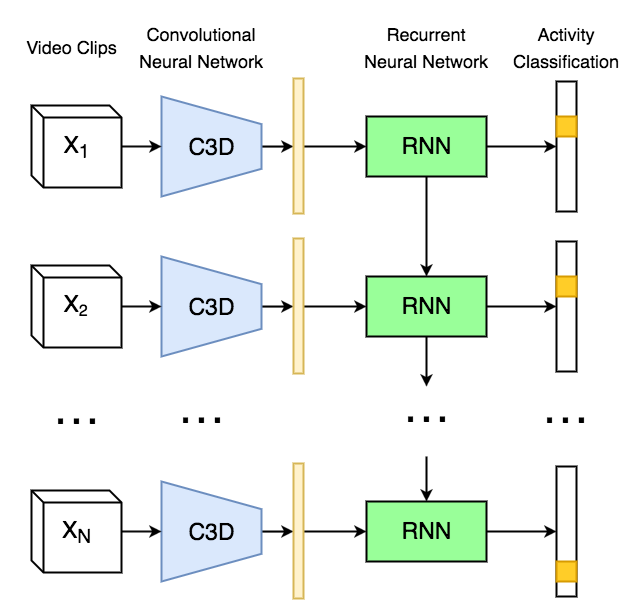}
\caption{Global architecture of the proposed pipeline.}
\label{fig:global_pipeline}
\end{figure}

We design a network that processes a sequence of C3D-f6 features from a video, and returns a sequence of class probabilities for each 16-frames clip.
We use LSTM layers, trained with dropout with probability $p = 0.5$ and a fully connected layer with a softmax activation. Figure~\ref{fig:global_pipeline} shows the proposed architecture.
Different configurations of the number of LSTM layers $N$ and the number of cells $c$ have been tested and are compared in Section~\ref{sec:resultats}.
Our proposed system has the following architecture: \texttt{input(4096) - dropout(0.5) - N $\times$ lstm(c) - dropout(.5) - softmax(K+1)} where $K$ is the number of activity classes at the dataset.

\subsection{Post-Processing}
Given a video, the prediction of our model is sequence of class probabilities for each 16-frame video clip. This output is post-processed to predict the activity class and temporally localize it.
First, to obtain the activity prediction for the whole video, we compute the average of the class probabilities over all video clips in the video.
We consider the class with maximum predicted probability as the predicted class.



To obtain the temporal localization of the predicted activity class, we first apply a mean filter of $k$ samples to the predicted sequence to smooth the values through time (see Equation~\ref{eq:smooth}).
Then, the probability of \textit{activity} (vs \textit{no activity}) is predicted for each 16-frames clip, being the \textit{activity} probability the sum of all probabilities of activity classes, and the \textit{no activity} probability, the one assigned to the background class.
Finally, only those clips with an \textit{activity} probability over a threshold $\gamma$ are kept and labeled with the previously predicted class.
Notice that, for each video, all predicted temporal detections are activity class.

\begin{equation}
	\tilde{p}_i(x) = \frac{1}{2k} \sum_{j=i-k}^{i+k} p_i(x)
    \label{eq:smooth}
\end{equation}



\section{Experiments}

\subsection{Dataset}
For all our experiments we use the dataset provided in the ActivityNet Challenge 2016~\cite{caba2015activitynet}. This dataset contains 640 hours of video and 64 million frames. The ActivityNet dataset is composed of untrimmed videos, providing temporal annotations for the given ground truth class labels. The dataset is split in 50\% for training, 25\% for validation and 25\% for testing.

\subsection{Training}
We train the network described in Section~\ref{sec:architecture} with the negative log likelihood loss, assigning a lower weight to background samples to deal with dataset imbalance (see Equation~\ref{eq:loss}).

\begin{equation}
    L(p,q) = - \sum_x \alpha(x) p(x) \log (q(x)), \text{ where } \alpha(x) =
    \begin{cases}
        \rho, & x = \text{background instance}\\
        1,    & \text{otherwise}
    \end{cases}
    \label{eq:loss}
\end{equation}

where $q$ is the predicted probability distribution and $p$ the ground truth probability distribution. In our experiments, we set $\rho = 0.3$.

The network was trained for 100 epochs, with a batch size of 256, where each sample in the minibatch is a sequence of 20 16-frame video clips. We use RMSprop~\cite{dauphin2015rmsprop} with a learning rate set to $10^{-5}$.


\subsection{Results}
\label{sec:resultats}

We evaluate our models using the metrics proposed in ActivityNet Challenge. For video classification, we use mean average precision (mAP) and Hit@3. For temporal localization, a prediction is marked correct only when it has the correct category and has IoU with ground truth instance larger than $0.5$, and mAP is used to evaluate the performance over the entire dataset.

\begin{table}[h]
    \parbox{.45\linewidth}{
        \centering
        \begin{tabular}{l|cc}
        \textbf{Architecture} & \textbf{mAP} & \textbf{Hit@3} \\
        \hline
        3 x 1024-LSTM & 0.5635 & 0.7437 \\
        2 x 512-LSTM & 0.5492 & 0.7364 \\
        1 x 512-LSTM & \bf0.5938 & \bf0.7576 \\
        \end{tabular}
        \vspace{.5cm}
        \caption{Results for classification task comparing different architectures.}
        \label{table:classification_by_architecture}
    }
    \hfill
    \parbox{.45\linewidth}{
        \centering
        \begin{tabular}{l|ccc}
        \textbf{$\gamma$} & \textbf{$k=0$} & \textbf{$k=5$} & \textbf{$k=10$} \\
        \hline
        0.2 & 0.20732 & \bf0.22513 & 0.22136 \\
        0.3 & 0.19854 & 0.22077 & 0.22100 \\
        0.5 & 0.19035 & 0.21937 & 0.21302 \\
        \end{tabular}
        \vspace{.2cm}
        \caption{mAP with an IoU threshold of $0.5$ comparing between values of $k$ and $\gamma$ on post-processing.}
        \label{table:detection_postprocessing_comparison}
    }
\end{table}

Table~\ref{table:classification_by_architecture} shows the performance of different network architectures. We tested configurations with different number of LSTM layers and different number of cells. These results indicate that all the networks presented high learning capacity over the data, but some over-fitting was observed with the deeper architectures, obtaining the best results with a single layer of 512-LSTM cells.

\begin{figure}[ht]
\centering
\includegraphics[width=1\linewidth]{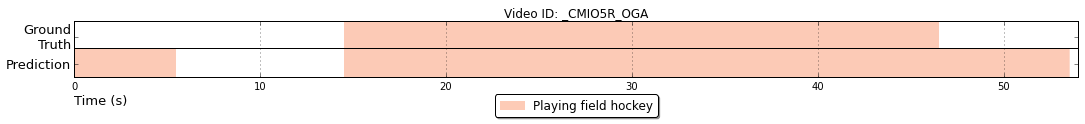}
\includegraphics[width=1\linewidth]{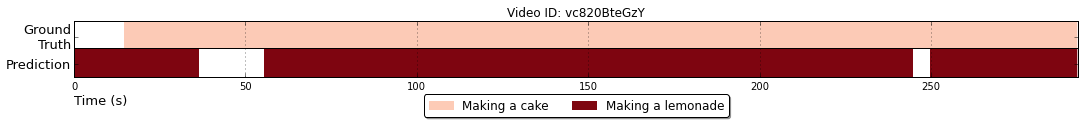}
\includegraphics[width=1\linewidth]{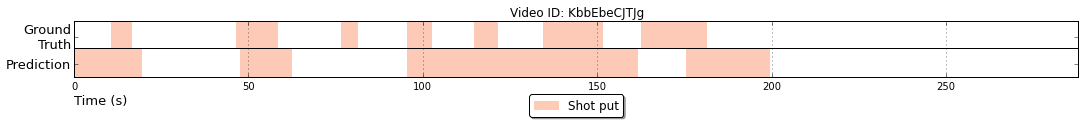}
\caption{Examples of temporal activity localization predictions.}
\label{fig:example_results_detect}
\end{figure}

\begin{figure}[h]
\centering
\begin{subfigure}[b]{.5\textwidth}
    \includegraphics[width=0.9\linewidth]{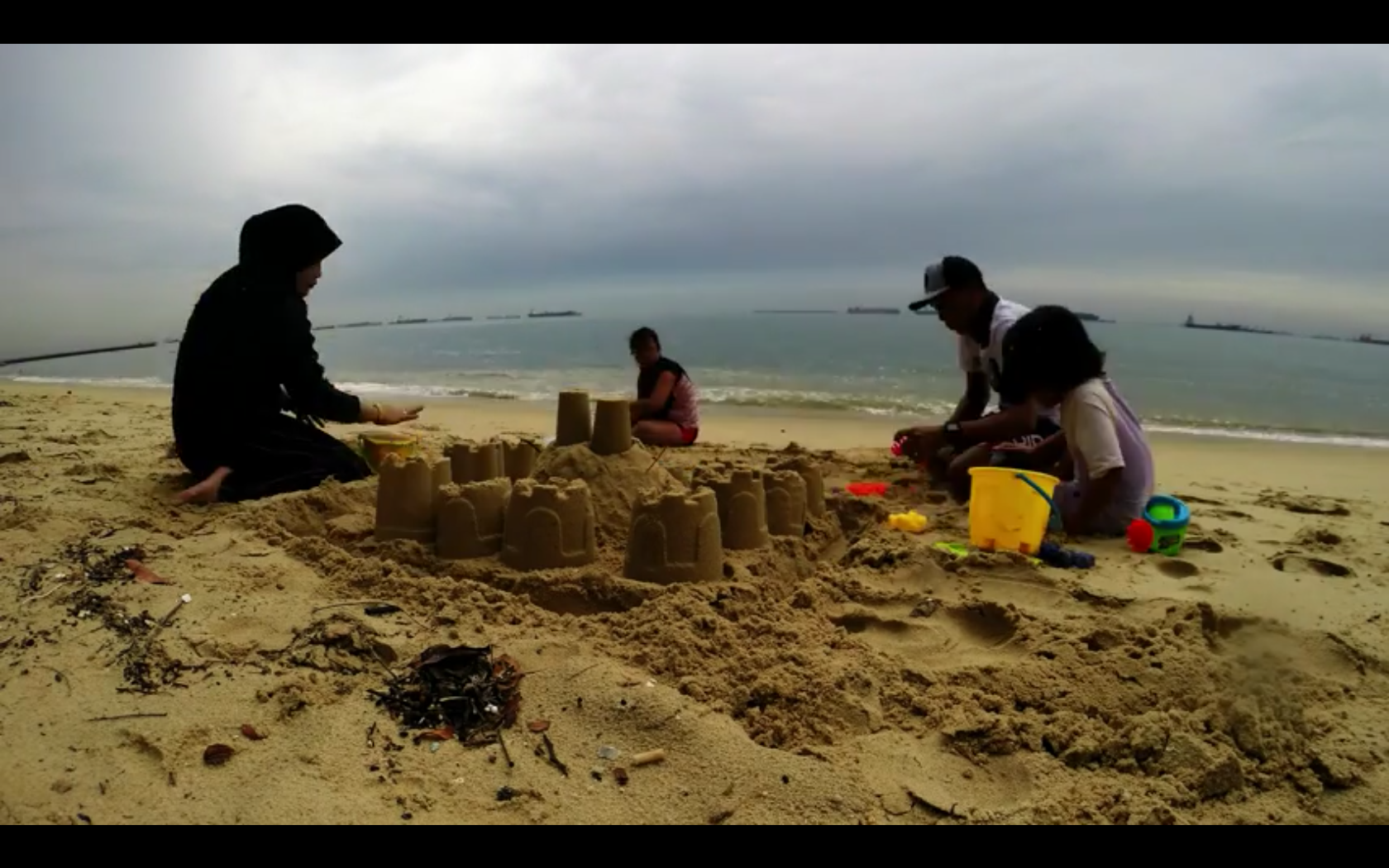}
    \texttt{Video ID: ArzhjEk4j\_Y \\
    Ground Truth: Building sandcastles \\
    \\
    Prediction: \\
    0.7896	Building sandcastles \\
    0.0073	Doing motocross \\
    0.0049	Beach soccer \\}
\end{subfigure}%
\begin{subfigure}[b]{.5\textwidth}
    \includegraphics[width=0.9\linewidth]{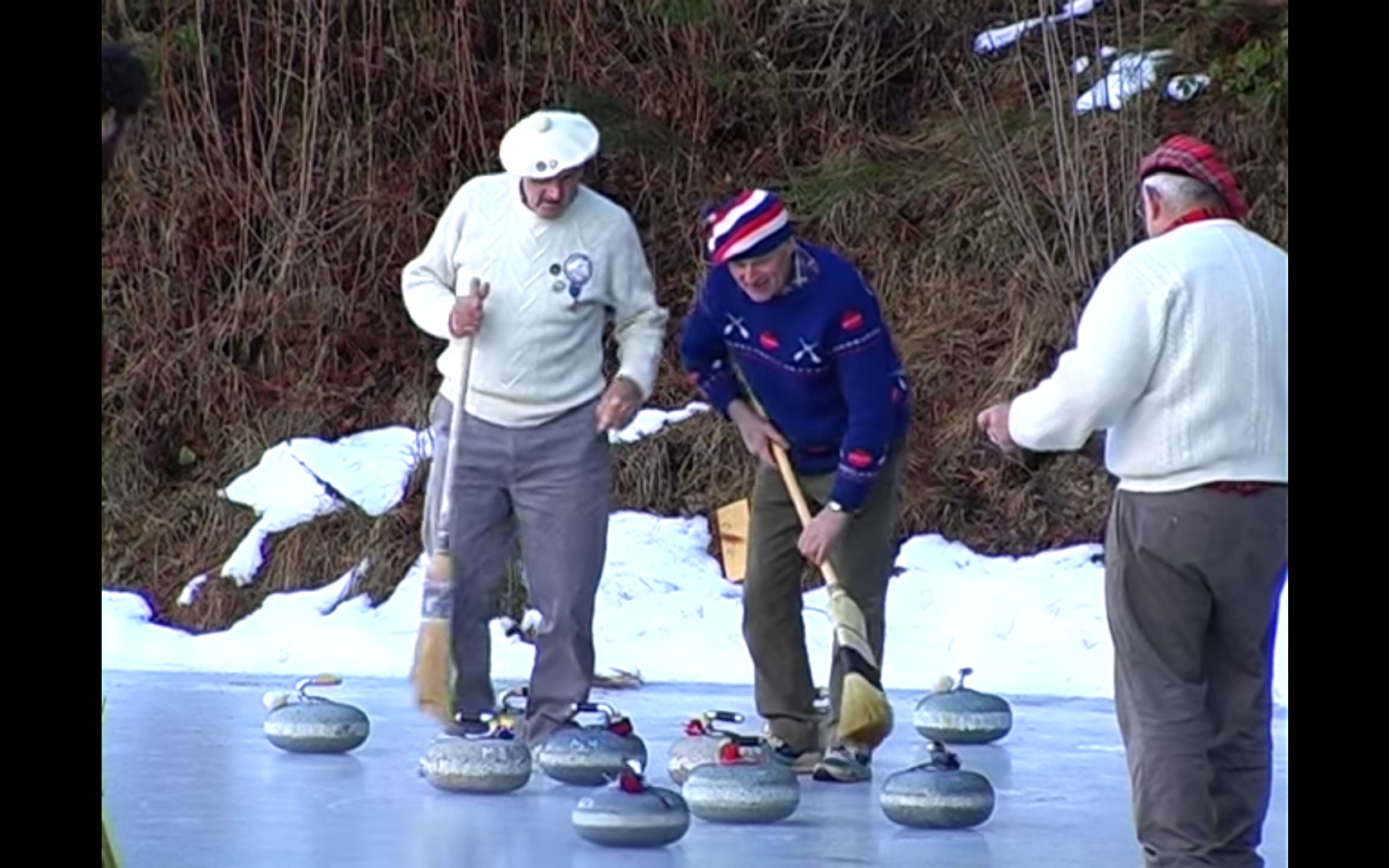}
    \texttt{Video ID: AimG8xzchfI \\
	Activity: Curling \\
    \\
    Prediction: \\
    0.3843	Shoveling snow \\
    0.1181	Ice fishing \\
    0.0633	Waterskiing \\}
\end{subfigure}

\caption{Examples of activity classification.}
\label{fig:example_results_classification}
\end{figure}

Fixing the architecture, we performed experiments for the temporal detection task using different values of $k$ and $\gamma$ in the post-processing stage.
Table~\ref{table:detection_postprocessing_comparison} shows results for the temporal activity localization task, where the effect of a mean smoothing filter can be seen, improving the localization performance.
Figures~\ref{fig:example_results_detect} and~\ref{fig:example_results_classification} show some examples of classification and temporal localization prediction for some instances of the dataset.

\section{Conclusion}

In this paper we propose a simple pipeline for both classification and temporal localization of activities in videos. Our system achieves competitive results on both tasks. The sequence to sequence nature of the proposed network offers flexibility to extend it to face more challenging tasks in video processing, e.g. where more than a single activity is present in the video. Future work will address end to end training of the model (3D-CNN + RNN), to learn better feature representations suitable for the dataset.


\section{Acknowledgements}

This work has been developed in the framework of the project BigGraph TEC2013-43935-R, funded by the Spanish Ministerio de Economia y Competitividad and the European Regional Development Fund (ERDF). The Image Processing Group at the UPC is a SGR14 Consolidated Research Group recognized and sponsored by the Catalan Government (Generalitat de Catalunya) through its AGAUR office. We gratefully acknowledge the support of NVIDIA Corporation with the donation of the GeForce GTX Titan Z used in this work.

\section*{}
{\small
\bibliographystyle{plainnat}
\bibliography{references}
}

\end{document}